\documentclass{article}

    \usepackage[preprint]{neurips_2025}

\usepackage[utf8]{inputenc} % allow utf-8 input
\usepackage[T1]{fontenc}    % use 8-bit T1 fonts
\usepackage{hyperref}       % hyperlinks
\usepackage{url}            % simple URL typesetting
\usepackage{booktabs}       % professional-quality tables
\usepackage{amsfonts}       % blackboard math symbols
\usepackage{nicefrac}       % compact symbols for 1/2, etc.
\usepackage{microtype}      % microtypography
\usepackage{xcolor}         % colors

\usepackage{wrapfig} 
\usepackage{times}
\usepackage{epsfig}
\usepackage{graphicx}
\usepackage{amsmath}
\usepackage{amssymb}
\usepackage{booktabs}
\usepackage{adjustbox}
\usepackage{multirow}
\usepackage{caption}
\usepackage{natbib}
\usepackage{xcolor}  
\usepackage{pifont}
\newcommand{\cmark}{\ding{51}} % 对号
\newcommand{\xmark}{\ding{55}} % 叉号
\usepackage{tcolorbox}
\title{PostAlign: Multimodal Grounding as a \\Corrective Lens for MLLMs}

\author{Yixuan Wu$^{1,2}$ \quad Yang Zhang$^{1,3}$ \quad Jian Wu$^2$ \quad Philip Torr$^{1}$ \quad Jindong Gu$^1$\\
{ $^1$University of Oxford \qquad $^2$Zhejiang University \qquad $^3$National University of Singapore}\\
{\tt yixuanwuwu7@gmail.com}
}
\vspace{2em}

\begin{document}

\maketitle

% \vspace{-2em}
\begin{abstract}
% \vspace{-0.5em}
Multimodal Large Language Models (MLLMs) excel in vision-language tasks, such as image captioning and visual question answering. However, they often suffer from over-reliance on spurious correlations, primarily due to linguistic priors that distract the model from leveraging actual visual information. To address these issues, we introduce MMGrounded-PostAlign, a post-multimodal alignment framework designed to enhance the visual understanding capabilities and mitigate the hallucinations of MLLMs. Our framework incorporates a multimodal grounding module for both visual grounding, which identifies the referred object in the image, and textual grounding, which generates the rationale for the final answer, ensuring that outputs are anchored in both visual and textual evidence. To mitigate the hallucinations, we introduce a negative rejection mechanism in the visual grounding module to distinguish grounded entities from non-existent objects influenced by linguistic biases. On the textual grounding side, we propose a selective reasoning mechanism that adjusts the model’s reasoning strategy based on query complexity. Extensive evaluations are conducted on benchmarks such as POPE, HaloQuest, VQAv2, MME, and MMBench showing significant improvements in fine-grained visual understanding and hallucination suppression. 
\end{abstract}

\section{Introduction}
Recently, the rapid development of Multimodal Large Language Models (MLLMs) has significantly advanced visual understanding by improving multimodal alignment between visual and textual representations. 
This multimodal alignment enables MLLMs to bridge the gap between modalities, leveraging large-scale vision encoders and pretrained language models to achieve remarkable results in tasks such as image captioning, visual question answering, and visually grounded dialogue~\cite{guo2025deepseek,team2024chameleon,wang2024qwen2,achiam2023gpt,liu2024improved,liu2023visual,team2023gemini}.
However, as tasks grow more demanding in terms of fine-grained visual understanding and complex reasoning, these models often fail to maintain robust alignment between modalities~\cite{liu2024survey,bai2024hallucination,zhou2023analyzing}, revealing critical limitations that hinder their robustness and reliability.

A key factor behind these limitations is the model’s reliance on spurious correlations~\cite{calude2017deluge,ye2024spurious} instead of genuine causal or contextual relationships. Rather than interpreting visual content based on meaningful interactions, MLLMs often default to statistical associations, such as objects frequently appearing together in images. This problem is further amplified by linguistic priors~\cite{leng2024mitigating}, which shape the model’s predictions based on common textual patterns rather than actual visual evidence. 

\begin{figure*}[t]
    \centering
    \includegraphics[width=0.96\textwidth]{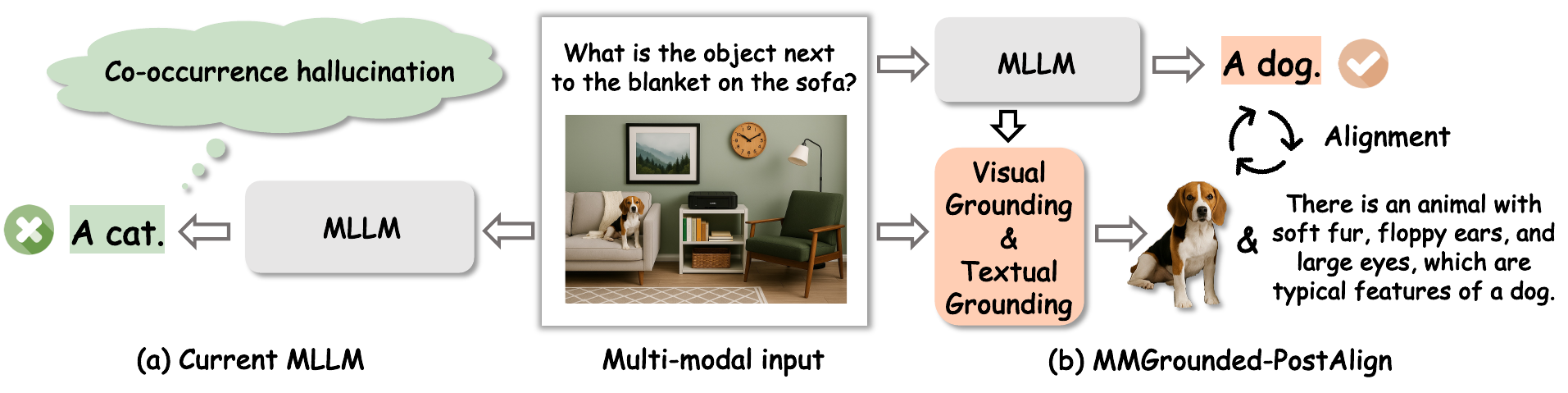}
    \caption{(a) Current MLLMs often struggle with spurious correlations, leading to co-occurrence hallucinations, such as incorrectly identifying a cat on the sofa. (b) To address this, we propose MMGrounded-PostAlign, which incorporates a multimodal grounding module that enhances answer accuracy by aligning and anchoring the final output with actual visual and textual evidence.}
    \label{fig:teaser}
\vspace{-1.7em}
\end{figure*}

One major issue caused by this bias is hallucination~\cite{liu2024survey,bai2024hallucination}, where the model generates content that does not exist in the image. Instead of relying on visual details, it selects words that are likely to appear together in text, even if they are incorrect in the given visual context. This reveals a lack of robust multimodal grounding.
Another challenge is the over-reliance on high-level visual cues. Instead of recognizing fine-grained object attributes, MLLMs often focus on general patterns like backgrounds or dominant colors. Linguistic priors encourage broad, text-driven generalizations that do not always align with the actual scene. As a result, the model struggles with precise visual understanding.
MLLMs also face difficulties in complex reasoning tasks. When interpreting object interactions, answering compositional questions, or inferring causal relationships, they often lack logical consistency. Rather than constructing well-grounded multimodal responses, the model may rely on familiar text patterns, leading to reasoning failures that prioritize language biases over actual visual input.

To address these challenges, MLLMs need stronger multimodal alignment to ensure that textual outputs remain faithful to visual input.
Specifically, we propose multimodal grounding as the corrective lens to mitigate these limitations and reduce reliance on spurious correlations. Multimodal grounding improves the alignment between modalities, helping the model establish more reliable connections between language and relevant sources of information.
Rather than relying on high-level correlations or co-occurrence-based embeddings, grounding can take various forms. It may link textual descriptions to visual regions, external knowledge bases, or contextual cues. By reinforcing consistency across modalities, grounding helps mitigate hallucinations, reducing the impact of spurious correlations and improving output reliability.

In this paper, we propose MMGrounded-PostAlign, a post-multimodal alignment framework designed to enhance the visual understanding capabilities of MLLMs. Our framework integrates a multimodal grounding module, which enables both visual grounding, identifying the referred object in the image, and textual grounding, which generates the rationale before the final answer. \textbf{This ensures that the MLLM’s outputs are firmly anchored in both real visual and textual evidence.}
To mitigate multimodal hallucinations, particularly those arising from linguistic biases, we introduce a negative rejection mechanism in the visual grounding module. This mechanism helps the model distinguish grounded entities from non-existent objects, reducing the risk of generating incorrect or hallucinated information influenced by language priors.
On the textual grounding side, we propose a selective reasoning mechanism that dynamically adapts the model’s reasoning strategy based on the complexity of the query. This mechanism enables the model to selectively incorporate textual reasoning when necessary, ensuring more contextually relevant responses across tasks of varying complexity levels.
By integrating these modules, our framework reinforces multimodal alignment, leading to more accurate visual understanding and reduced hallucinations in MLLMs.

Our contributions are summarized as follows:
\vspace{-0.3em}
\begin{itemize}
\item We propose MMGrounded-PostAlign, a post-multimodal alignment framework that enhances visual understanding in MLLMs by aligning visual and textual modalities through multimodal grounding, ensuring outputs are anchored to actual visual and textual evidence.
% \vspace{-0.1em}
\item We introduce a negative rejection mechanism in visual grounding to mitigate hallucinations, allowing the model to differentiate between grounded objects and hallucinated artifacts.
% \vspace{-0.1em}
\item In textual grounding, we propose a selective reasoning mechanism that adapts the model's reasoning strategy based on the complexity of the input query, improving the model's ability to handle reasoning tasks with various complex levels.
% \vspace{-0.1em}
\item We evaluate our method on benchmarks such as HaloQuest, POPE, VQAv2, MMBench, MME, and RefCOCO, achieving significant improvements in visual understanding and hallucination suppression, while preserving the general reasoning capabilities of MLLMs. 
\end{itemize}

\section{Related Work}
\subsection{Vision-Language Grounding Models}
Grounding in vision-language models involves establishing correspondences between textual descriptions and image regions~\cite{xie2023described}. 
Early grounding methods~\cite{yu2018mattnet,yang2019fast} primarily adopted object detection pipelines, while later models~\cite{kamath2021mdetr,yao2022detclip,li2022grounded,liu2024grounding} integrated vision-language pretraining for better end-to-end performance. 
Beyond bounding box grounding, segmentation-based grounding~\cite{wu2020phrasecut,wu2023advancing,wu2022language} addressed coarse granularity by enabling pixel-wise segmentation for finer localization.

With the rise of MLLMs, grounding tasks have been integrated into generalist frameworks. Some models, like Kosmos-2~\cite{peng2023kosmos} and Shikra~\cite{chen2023shikra}, formulate grounding as text-based bounding box prediction, while others~\cite{zhang2024llava,zhao2023bubogpt,lai2024lisa,xia2024gsva,rasheed2024glamm,ren2024pixellm,zhang2024groundhog}, integrate separate grounding modules into MLLMs. 
These methods primarily focus on enhancing grounding accuracy within a single-task framework, typically using MLLMs to control or guide grounding.
\textbf{In contrast, our work shifts this interaction: rather than using MLLMs for grounding tasks, we leverage multimodal grounding to enhance MLLMs' visual understanding and reduce hallucinations from linguistic biases.}

\subsection{Hallucination in MLLMs}
Hallucination in MLLMs, characterized by contradictions between image input and textual output, has been a prevalent issue~\cite{liu2024survey,bai2024hallucination}. To alleviate hallucinated content, existing works can be divided into the following two directions.  

The first focuses on post-processing approaches, including post-hoc corrections~\cite{zhou2023analyzing,yin2024woodpecker,lee2023volcano,zhou2023analyzing} and specialized decoding~\cite{huang2024opera,chen2024halc,leng2024mitigating,zhu2024ibd,zhao2024mitigating,deng2024seeing,wang2024mllm}. For example, Volcano~\cite{lee2023volcano} introduces a self-revising mechanism to
reduce hallucination.
% ; while VCD~\cite{leng2024mitigating} aims to reduce the over-reliance on language priors by calibrating the decoding probability distribution in a contrastive manner.
However, these methods often require increased inference time, limiting their generalizability and scalability across diverse data domains and model sizes.  

The second line of work focuses on training-based methods. Some of them focus on data-level improvements, including the introduction of negative data~\cite{liu2023mitigating}, counterfactual data~\cite{yu2024hallucidoctor}, and reducing noise in existing datasets~\cite{wang2024mitigating,yue2024less}. Reinforcement learning (RL) has also been explored to guide MLLMs toward hallucination-free outputs~\cite{zhao2023beyond,li2023silkie,gunjal2024detecting,sun2023aligning}. Instead of solely relying on data augmentation or reinforcement learning, our method integrates a multimodal grounding module for stronger multimodal alignment, ensuring that model outputs are firmly anchored in real evidence.

\section{Method}
\label{sec:method}
In this section, we present our post-multimodal alignment framework, MMGrounded-PostAlign, with the integration of both visual and textual grounding. 
In this section, we introduce MMGrounded-PostAlign, a post-multimodal alignment framework that integrates both visual and textual grounding. 
This multimodal grounding acts as the corrective lens, reducing the MLLM's over-reliance on linguistic priors, while simultaneously enhancing its visiual understanding capabilities. 
Specifically, it (1) provides visually-grounded cues to guide MLLM outputs, and (2) anchors the reasoning process with textual grounding, ensuring that outputs are supported by interpretable rationales.
\subsection{Framework Overview}
\begin{figure*}[t]
    \centering
    \includegraphics[width=0.96\textwidth]{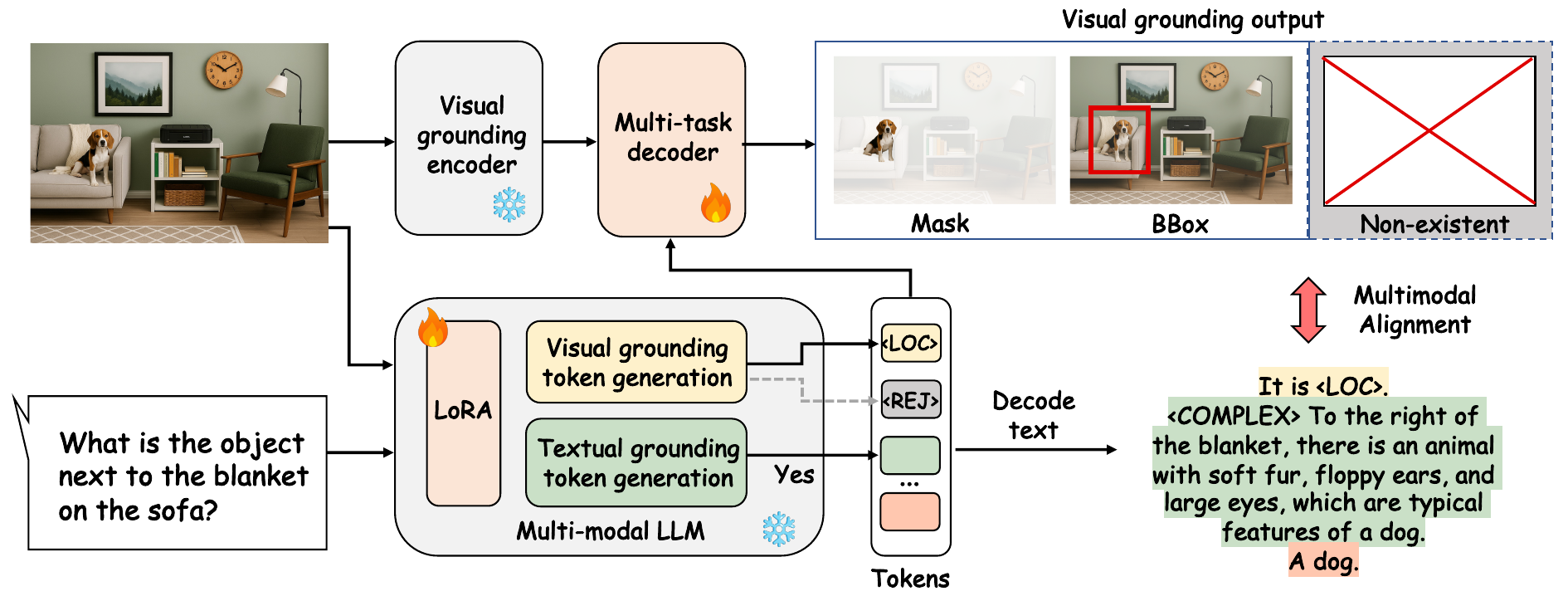}
    \caption{The pipeline of our proposed MMGrounded-PostAlign. Given an image and text query, the MLLM generates the tokens of visual grounding (\texttt{<LOC>}), textual grounding, and the final answer.
    The last-layer embedding of \texttt{<LOC>} is fed to multi-task decoder and then decoded into a segmentation mask and bounding box for the target object. 
    When no target object exists in the image, the visual grounding token is replaced by \texttt{<REJ>}, which is directly assigned an empty mask and bounding box in the multi-task decoder. The textual grounding facilitates the generation of a rationale prior to the final answer for complex queries.}
    \label{fig:main}
\vspace{-1.7em}
\end{figure*}

As shown in Figure~\ref{fig:main}, our framework comprises three key components:
(1) A MLLM that processes image and text inputs to predict the tokens of visual grounding, textual grounding, and the final answer.
(2) A visual grounding encoder that extracts image features.
(3) A multi-task decoder that performs visual grounding of segmentation mask and detection bounding box, using the visual grounding token as prompts.

\vspace{-2mm}
\paragraph{Multi-modal Input and Structured Output.}
Given an image $\mathcal{I}$ and a text query $\mathcal{Q}$, the MLLM is tasked with generating a structured output $\mathcal{A}$, as:
\begin{equation}
    \mathcal{A} = \text{MLLM}(\mathcal{I}, \mathcal{Q}),
\end{equation}
\begin{equation}
    \mathcal{A} = \{\mathcal{V}, \mathcal{T}, \mathcal{F}\},
\end{equation}
where $\mathcal{V}$ denotes the visual grounding token, $\mathcal{T}$ represents the textual grounding token for rationale, and $\mathcal{F}$ is the final answer token.

\vspace{-2mm}
\paragraph{Visual Grounding Prediction.}
The last-layer embedding of the visual grounding token $\texttt{<LOC>}$ is extracted from the MLLM and passed through an MLP projection layer to obtain the prompt embeddings for the subsequent visual grounding. Meanwhile, the visual grounding encoder extracts dense visual features from the input image. These dense visual features, along with the prompt embeddings, are fed into the multi-task decoder, which generates the final visual grounding outputs. 
\vspace{-2mm}
\paragraph{Grounding-Augmented Final Answer Generation.}
The visual grounding outputs and textual grounding rationales act as implicit constraints that guide the MLLM’s final answer generation. 
The textual grounding provides a structured framework for generating interpretable rationales, which, when integrated with the visual grounding cues, anchor the model’s reasoning in both visual and linguistic contexts.
This guidance helps mitigate the model's over-reliance on linguistic priors, enhancing its ability to perceive and understand visual information more accurately.

\subsection{Rejecting Non-existent Objects}
Despite the advanced understanding capabilities of MLLMs, they are often prone to hallucinating non-existent objects due to strong linguistic priors and co-occurrence biases present in the training data. These hallucinations manifest in various forms, including incorrect attributes, inaccurate spatial locations, or entirely fabricated objects. 

To mitigate this, we introduce a negative rejection mechanism, which enables the model to explicitly reject non-existent objects rather than hallucinating their presence. Specifically, if the referents are absent from the image, we enforce the MLLM to predict a special $\texttt{<REJ>}$ token to replace the corresponding $\texttt{<LOC>}$ token. 
In the multi-task decoder, the predicted $\texttt{<REJ>}$ tokens are directly assigned empty masks and bounding boxes, effectively bypassing the decoding process. The primary advantage of the negative rejection mechanism is its ability to mitigate the model's over-reliance on linguistic priors. Many MLLMs tend to associate high-frequency co-occurring objects (e.g., predicting a ``\textit{table}" when a ``\textit{chair}" is present) and produce incorrect outputs. By explicitly learning to reject such negative samples, the model becomes more capable of distinguishing between genuine visual grounding and misleading language priors.
Additionally, incorrect attributes and spatial misplacement, such as predicting a ``\textit{red apple}" when only green apples are present, are also suppressed, as the model is penalized for assigning masks or bounding boxes to absent entities.
% \vspace{-2mm}
\paragraph{Negative Rejection Loss.}  
To further reinforce the model’s ability to reject hallucinated objects, we introduce a negative rejection loss, which explicitly penalizes incorrect grounding predictions when the referent is absent. The loss is defined as:
\begin{equation}
\mathcal{L}_{\text{rej}} = -\frac{1}{N} \sum_{i=1}^N \left[ y_i^{\text{rej}} \log p_i^{\text{rej}} + (1 - y_i^{\text{rej}}) \log (1 - p_i^{\text{rej}}) \right],
\end{equation}
where \( y_i^{\text{rej}} \in \{0,1\} \) denotes whether the referent in sample \( i \) should be rejected.

\subsection{Selective Reasoning in Textual Grounding}
\label{sec:reasoning}
Previous work~\cite{zhang2023multimodal} has shown that generating intermediate rationales through textual grounding improves the reasoning capabilities of MLLMs, leading to improved answer accuracy for complex tasks. 
In our multimodal grounding-assisted MLLM framework, we argue that not all queries require explicit rationale generation via textual grounding. Instead, a selective approach is needed to balance the visual output from the visual grounding module with the textual answer generation of the MLLM. Our experiments, as shown in Table~\ref{tab:reasonseg}, also support this hypothesis.
To this end, we propose a selective reasoning mechanism, which allows the MLLM to determine whether explicit contextual reasoning is needed.
Specifically, during training, we categorize queries into \texttt{<SIMPLE>} and \texttt{<COMPLEX>} types:
\begin{itemize}
  \item For simple queries, the MLLM directly outputs the visual grounding token and the final answer, skipping rationale generation. For example, for the simple query: \textit{“User: \texttt{<IMAGE>} What color is the car in this image?” “Assistant: It is \texttt{<LOC>}. \texttt{<SIMPLE>}. The car is red.”}
 \item For complex queries, the MLLM's output includes the visual grounding token, the contextual rationale, and the final answer. For example, for the complex query: \textit{“USER: \texttt{<IMAGE>} Which food in the picture contains the most protein?” “ASSISTANT: It is \texttt{<LOC>}. \texttt{<COMPLEX>}. This image contains oranges, eggs, vegetables, and buns. Among them, eggs are the richest in protein. ”}  
\end{itemize}
% \vspace{1mm}
To enable automatic complexity assessment during inference, we incorporate self-reflection prompting into the user instructions: 

% \vspace{2mm}
\noindent
\begin{minipage}{\textwidth}
\begin{tcolorbox} 
\small
\textit{Given an image and a text query, first assess whether answering the query requires a rationale. If the answer is directly observable from the image without additional reasoning, classify it as \texttt{<SIMPLE>} and provide only the final answer. If answering requires logical inference or contextual understanding, classify it as \texttt{<COMPLEX>} and generate a rationale before providing the final answer.}
\end{tcolorbox}
% \vspace{1mm}
\end{minipage}

\paragraph{Selective Reasoning Loss.}  
We also design a selective reasoning loss to supervise the model in identifying whether a query requires complex reasoning. Each query is classified as either \texttt{<SIMPLE>} or \texttt{<COMPLEX>}, and the loss is computed as:
\begin{equation}
\mathcal{L}_{\text{reason}} = -\frac{1}{N} \sum_{i=1}^N \left[ y_i^{\text{rea}} \log p_i^{\text{rea}} + (1 - y_i^{\text{rea}}) \log(1 - p_i^{\text{rea}}) \right],
\end{equation}
where \( y_i^{\text{rea}} \in \{0,1\} \) indicates whether query \( i \) requires generating a rationale.

\subsection{Training Objective}
In summary, we finetune the MLLMs using LoRA~\cite{hu2022lora} while jointly optimizing the multi-task decoder. The overall loss function is defined as:
\begin{equation}
    \mathcal{L} = \lambda_1 \mathcal{L}_{\text{rej}}  + \lambda_2 \mathcal{L}_{\text{reason}} +  \mathcal{L}_{\text{ground}} +  \mathcal{L}_{\text{text}},
\end{equation}
where the visual grounding loss $\mathcal{L}_{\text{ground}}$ consists of a joint optimization of detection loss $\mathcal{L}_{\text{det}}$ and segmentation loss $\mathcal{L}_{\text{seg}}$, formulated as:
\begin{equation}
    \mathcal{L}_{\text{det}} = \mathcal{L}_{\text{smooth-L1}}(\hat{y}_{\text{bbox}}, y_{\text{bbox}}) + \mathcal{L}_{\text{GIoU}}(\hat{y}_{\text{bbox}}, y_{\text{bbox}}),
\end{equation}
\begin{equation}
\mathcal{L}_{\text{seg}} = \mathcal{L}_{\text{BCE}}(\hat{y}_{\text{mask}}, y_{\text{mask}}) +  \mathcal{L}_{\text{DICE}}(\hat{y}_{\text{mask}}, y_{\text{mask}}),
\end{equation}
and $\mathcal{L}_{\text{text}}$ is the cross entropy language modeling loss, defined as:
\begin{equation}
  \mathcal{L}_{\text{text}} = \mathcal{L}_{\text{LM}}(\hat{y}_{\text{txt}}, y_{\text{txt}}).
\end{equation}

\section{Experiment}
\label{sec:exp_all}
\subsection{Setups} 
\label{sec:exp}
\noindent\textbf{Training Data Formulation.}
We construct a diverse multimodal training dataset, where each instance is annotated with a reasoning-type token: \texttt{<SIMPLE>} or \texttt{<COMPLEX>}. The \texttt{<SIMPLE>} label denotes low-complexity queries requiring minimal reasoning, while \texttt{<COMPLEX>} is used for tasks involving higher-order or indirect reasoning. Additionally, negative samples are included with the \texttt{<REJ>} token, indicating the absence of a visual referent. 
% Detailed data construction and labeling procedures are provided in the \textit{Appendices}.

\noindent\textbf{Network Architecture.}  
Our framework is built upon LLaVA-1.5-7B and LLaVA-1.5-13B~\cite{liu2024improved} as the MLLM backbone, and ViT-H SAM~\cite{kirillov2023segment} as the visual grounding backbone. To preserve the pre-trained MLLM’s knowledge, we employ LoRA~\cite{hu2022lora} for parameter-efficient fine-tuning, while freezing the visual grounding encoder. 
The multi-task decoder, along with the LLM token embeddings, LLM head, and projection layer, is fully fine-tuned. The decoder includes both the mask and bounding box decoders. The hidden embeddings of visual grounding tokens (e.g., \texttt{<LOC>}) serve as prompts for SAM, conditioning its mask decoder to predict object masks, while a lightweight MLP regressor predicts bounding box coordinates from SAM-extracted features.

\noindent\textbf{Implementation Details.} 
Training is conducted using the DeepSpeed engine for efficient large-scale optimization. We employ the AdamW~\cite{loshchilov2017decoupled} optimizer with a learning rate of 0.0003 and no weight decay. The learning rate follows a WarmupDecayLR schedule with 100 warmup iterations. 
We use a per-device batch size of 2 with a gradient accumulation step of 10.
% For detailed hyperparameter settings, please refer to the \textit{Appendices}.

\subsection{Main Results and Analysis}
In this section, we present results across three benchmark categories: (1) hallucination datasets (HaloQuest~\cite{wang2024haloquest} and POPE~\cite{li2023evaluating}), (2) generalization and reasoning datasets (VQAv2~\cite{goyal2017making}, MMBench~\cite{liu2024mmbench}, MME~\cite{liang2024survey}), and (3) grounding tasks (RefCOCO series~\cite{kazemzadeh2014referitgame} and ReasonSeg~\cite{lai2024lisa}). 
Our method demonstrates that, while preserving the general reasoning capabilities of MLLMs, it not only mitigates hallucinations but also enhances visual understanding.
We then highlight five key findings that identify the primary sources of hallucinations and demonstrate how our method addresses them.

\begin{figure*}[t]
    \centering
    \includegraphics[width=\textwidth]{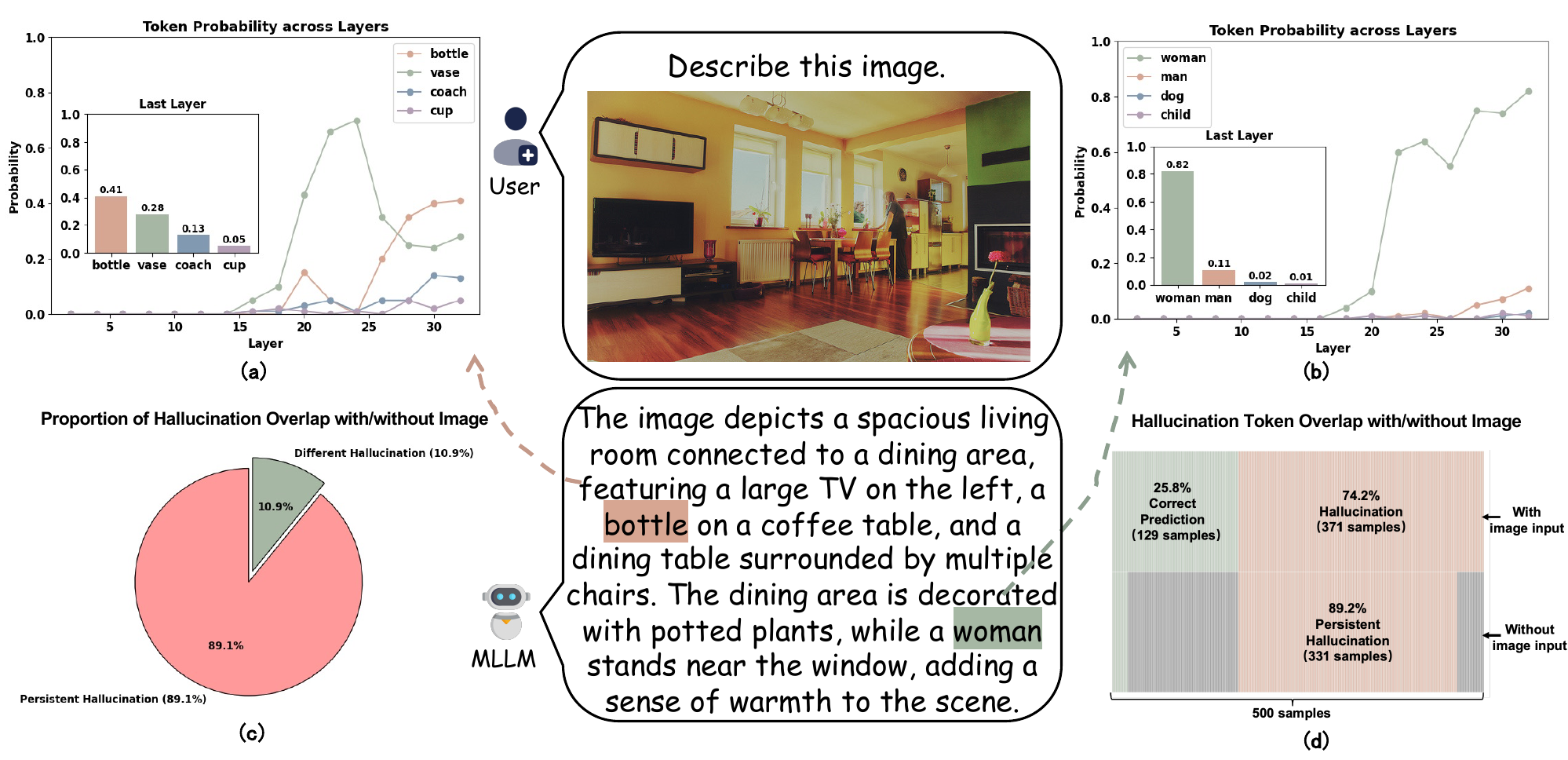}
    \caption{(a)(b) Token probability distributions across transformer layers, showing distinct trends for hallucinated (pink) and non-hallucinated (green) tokens. (c)(d) Removing visual input yields an $89.2\%$ overlap in hallucinated tokens, supporting the hypothesis that hallucinations arise from linguistic biases rather than visual misinterpretation.}
    \label{fig:prior}
\vspace{-2em}
\end{figure*}
\noindent \textbf{Finding 1: Linguistic Priors Override Visual Information in MLLMs.}
As illustrated in Figure~\ref{fig:prior}, we investigate the extent to which MLLMs rely on linguistic priors over visual inputs. Our analysis focuses on the layer-wise probability distributions of generated tokens during decoding in (a)-(b), and measures how the presence of visual input influences token generation in (c)-(d).
% (see \textit{Appendices} for setup details).

In (a) and (b), we analyze the Top-4 tokens ranked by probability in the final decoding layer. Non-hallucinated tokens such as ``woman" exhibit high probabilities from the 20th layer, whereas hallucinated tokens like ``bottle" only reach a higher probability around the 30th layer. 
Notably, in (a), the probability of the ground-truth token ``vase" sharply declines after the 25th layer, eventually falling below hallucinated tokens. 
We hypothesize that linguistic priors progressively suppress visual information during decoding, shifting the model toward language-driven generation, even against visual evidence.

In (c) and (d), we conducted a statistical analysis to further validate our hypothesis by removing the image input and analyzing whether hallucinated tokens continue to appear. Our results show that across 500 randomly sampled images from MSCOCO, the hallucinated token overlap rate reaches $89.2\%$.
In (d), we further dissect the token overlap rate. The upper portion of the figure illustrates that, with image input, hallucinations occur in $74.2\%$ of all the samples (highlighted in pink). Among these hallucinated samples, when the image input is removed, as shown in the lower portion of the figure (the pink section), $89.2\%$ of them still exhibit persistent hallucinations. This shows that, regardless of the presence of image input, a significant portion of hallucinations persists. This confirms that linguistic priors dominate hallucination generation, often overriding visual evidence.

\noindent \textbf{Finding 2: Explicit Visual Grounding Mitigates Hallucinations.}
To investigate the extent to which the visual grounding mechanism enhances MLLM's visual understanding and mitigates visual hallucinations, we first conduct an ablation study on the HaloQuest benchmark.
As shown in Table~\ref{tab:halo}, the first row presents the baseline where the visual grounding module is entirely removed. 
The results indicate that the baseline exhibits poor performance, particularly in the \textit{False Premise} and \textit{Insufficient Context} categories.
Introducing the mask grounding (denoted as \texttt{<SEG>}) and the bounding box grounding (denoted as \texttt{<DET>}) significantly improves performance. 
Furthermore, when both mask grounding and bounding box grounding are incorporated, the MLLM achieves even better results. 
Notably, the introduction of the \texttt{<REJ>} token for negative rejection further suppresses hallucinations, especially in the \textit{False Premise} and \textit{Insufficient Context} categories. This is primarily because these categories contain a substantial number of anti-concept scenarios. For example, a question such as \textit{``What breed of dog is in the image?''} may have a ground truth response of \textit{``There is no dog in the image.''}, which directly evaluates the MLLM’s ability to reject misleading premises.

Additionally, as shown in Table~\ref{tab:pope}, we further investigate different grounding strategies. A common approach to integrating MLLM with grounding mechanisms is treating \textbf{B}ounding boxes as language \textbf{T}okens for unified \textbf{L}earning (BTL). Based on this strategy, we explore two training paradigms:
\textbf{(1) BTL-Generation:} The input consists of an image and a referring text, while the MLLM directly generates the bounding box coordinates.
\textbf{(2) BTL-Caption:} The input is an image, and the output is a caption describing the image content, where the target object's bounding box is embedded within the generated text.
Here, {\textbf{baseline}} refers to our proposed framework with the visual grounding module removed, while retaining other designs (i.e., selective reasoning).
The results on the POPE hallucination benchmark indicate that incorporating BTL-Generation does not yield performance gains, while BTL-Caption provides moderate improvement. Combining both approaches further enhances performance.  
However, our proposed explicit visual grounding module significantly outperforms these methods. By enforcing a higher degree of multimodal alignment and effectively rejecting negative samples, our method substantially reduces object and attribute hallucinations.

\begin{table*}[t]
\centering
\footnotesize
\vspace{-2em}
\setlength\tabcolsep{0.06cm}
\caption{An ablation study evaluates the impact of visual grounding tokens (\texttt{<LOC>}) on the HaloQuest benchmark~\cite{wang2024haloquest}. Here, \texttt{<SEG>} and \texttt{<DET>} are specialized types of \texttt{<LOC>} tokens, where \texttt{<SEG>} generates segmentation masks and \texttt{<DET>} produces bounding boxes. The results show that combining \texttt{<SEG>} and \texttt{<DET>} enhances visual understanding, while \texttt{<REJ>} effectively suppresses hallucinations in misleading or insufficient context scenarios.}
\begin{tabular}{ccc|cc|cc|cc}
\toprule
\multicolumn{3}{c}{{Method}}  & \multicolumn{2}{c}{{False Premise}} & \multicolumn{2}{c}{{Visually Challenging}} & \multicolumn{2}{c}{{Insufficient Context}} \\ 
\cmidrule(lr){1-3} \cmidrule(lr){4-5} \cmidrule(lr){6-7} \cmidrule(lr){8-9}
{\texttt{<SEG>}} & {\texttt{<DET>}} & {\texttt{<REJ>}}  &  {Human Eval} & {Auto-Eval} & {Human Eval} & {Auto-Eval} & {Human Eval} & {Auto-Eval} \\ 
\midrule
% \cmidrule(lr){1-3} \cmidrule(lr){4-5} \cmidrule(lr){6-7} \cmidrule(lr){8-9}
\xmark & \xmark & \xmark & 2.0 & 2.3 & 23.5 & 23.0 & 2.5 & 1.7 \\
\cmark & \xmark & \xmark & 6.5 & 7.2 & 30.1 & 30.6 & 7.4 & 8.2 \\
\xmark & \cmark & \xmark & 8.2 & 8.9 & 31.1 & 31.7 & 6.6 & 7.4 \\
\cmark & \cmark & \xmark & 9.9 & 10.5 & 33.9 & 35.0 & 9.9 & 11.6 \\
\cmark & \cmark & \cmark & \textbf{33.2} & \textbf{33.9 }& \textbf{38.3} & \textbf{37.2} & \textbf{31.4} & \textbf{32.2} \\
\bottomrule
\end{tabular}
\label{tab:halo}
\vspace{-1.3em}
\end{table*}

\begin{table*}[t]
    \centering
    \footnotesize
    \vspace{-1em}
    \setlength\tabcolsep{0.23cm}
    \caption{An ablation study examines grounding strategies on the POPE, MME, MMBench, VQAv2 benchmark. Results show that explicit visual grounding module outperforms BTL-based methods by enforcing stronger multimodal alignment.}
    \begin{tabular}{lccccccc}
        \toprule
        & \multicolumn{3}{c}{{POPE~\cite{li2023evaluating}}} 
    & {MME} 
    & \multicolumn{2}{c}{{MMBench~\cite{liu2024mmbench}}} &VQAv2\\
        & {Ran} & {Pop} & {Adv} & {\cite{liang2024survey}} & \multicolumn{1}{c}{{EN}} & {CN} & {\cite{goyal2017making}} \\
        % \midrule
        \cmidrule(lr){1-1} \cmidrule(lr){2-4} \cmidrule(lr){5-5} \cmidrule(lr){6-7}  \cmidrule(lr){8-8} 
        Baseline-7B   &83.3 &80.1 &78.2 &1504.6 &62.2 &57.7 &77.4 \\
         + BTL-Generation &82.7&80.3&79.2 &1489.4  &59.2 &54.2 &75.9 \\   
         + BTL-Caption &84.5&81.0&79.9 & 1499.6&59.7 &54.5 &76.6 \\  
         + BTL-Generation + BTL-Caption &84.9&81.9&80.6 &1505.4 &60.3 &57.4 &76.2 \\   
        % \midrule
                \cmidrule(lr){1-1} \cmidrule(lr){2-4} \cmidrule(lr){5-5} \cmidrule(lr){6-7}  \cmidrule(lr){8-8} 
        MMGrounded-PostAlign-7B    &\textbf{86.6}&\textbf{84.2}&\textbf{82.3} &\textbf{1514.3} &\textbf{63.9} &\textbf{58.7} &\textbf{78.8} \\
        % \midrule
                \cmidrule(lr){1-1} \cmidrule(lr){2-4} \cmidrule(lr){5-5} \cmidrule(lr){6-7}  \cmidrule(lr){8-8} 
        Baseline-13B   &85.4 &82.2 &79.2 &\textbf{1520.3}&66.8 &62.2 &79.1 \\
         + BTL-Generation &84.4&80.9&78.3 &1501.7 &65.9 &62.4 &78.0 \\   
         + BTL-Caption &85.1&81.7&78.8 &1509.2 &65.2 &61.4 &79.4 \\  
         + BTL-Generation + BTL-Caption &86.8&83.4&81.7 &1504.9 &65.4 &62.2 &79.2 \\  
        % \midrule
                \cmidrule(lr){1-1} \cmidrule(lr){2-4} \cmidrule(lr){5-5} \cmidrule(lr){6-7}  \cmidrule(lr){8-8} 
        MMGrounded-PostAlign-13B    &\textbf{88.9}&\textbf{87.3}&\textbf{85.6} &1517.4 &\textbf{68.9} &\textbf{63.2} &\textbf{79.9} \\
        \bottomrule
        \end{tabular}
    \label{tab:pope}
    \vspace{-1.3em}
\end{table*}

\noindent \textbf{Finding 3: Our MLLM Retains Strong Reasoning and Generalization Abilities.}
In Table~\ref{tab:pope}, we further investigate whether explicitly incorporating visual outputs affects the general reasoning capabilities of MLLMs. To evaluate this, we test our method on several widely-used datasets for generalization and reasoning, including MME, MMBench, and VQAv2.
Our results reveal that our post-alignment method does not degrade the model's reasoning ability compared to the baseline MLLM. In contrast, we observe that the BTL-Generation training paradigm results in a notable decline in the baseline MLLM's reasoning ability. This suggests that the approach used in BTL-Generation may inadvertently bias the model towards overfitting to visual bounding box information at the cost of abstract reasoning, leading to a drop in generalization performance.

\noindent \textbf{Finding 4: Selective Reasoning in Textual Grounding Optimizes Efficiency and Accuracy.}
% Previous studies indicate that hallucinations often stem from inadequate reasoning on complex queries. 
Previous work~\cite{zhang2023multimodal} has shown that generating intermediate rationales through textual grounding improves the reasoning capabilities of MLLMs, leading to improved answer accuracy for complex tasks. 
In Table~\ref{tab:reasonseg}, we compare three textual grounding strategies:  
\textbf{(1) Pre-Reasoning:} A separate MLLM (the same as main MLLM) generates a rationale, which is then fed to the main MLLM for final answer generation.
\textbf{(2) Inter-Reasoning:} The main MLLM integrates reasoning, simultaneously generating rationale and final answers for each query.  
\textbf{(3) Selective-Reasoning:} Dynamically assesses query complexity and generates rationale only for complex queries. 
Here, \textbf{baseline} refers to our proposed framework with the selective reasoning strategy removed, while retaining the visual grounding module. 
Experimental results indicate that pre-reasoning strategy achieves better performance than inter-reasoning strategy; However, it involves multiple inference rounds, which reduces overall efficiency.
Our proposed selective-reasoning approach not only maintains strong performance but also operates with a single inference round.
Also, we augment the original benchmark queries into three difficulty levels (easy, medium, hard) to study the relationship between reasoning strategies and query complexity.
% (see \textit{Appendices} for details).
We observe that for simple queries, excessive reasoning can slightly degrade performance due to overthinking. Conversely, for complex queries, explicit reasoning is necessary to ensure accurate predictions.  

\begin{table}[t]
\centering
\caption{An ablation study on ReasonSeg~\cite{lai2024lisa} shows that selective reasoning in textual grounding performs best by adapting to different query complexity, which avoids overthinking on simple queries and ensures sufficient reasoning for complex ones.}
    \footnotesize
    \vspace{0.1em}
    \setlength\tabcolsep{0.23cm}
\begin{tabular}{lcccccc}
\toprule
\multirow{2}{*}{Method} & \multicolumn{2}{c}{{Easy}}& \multicolumn{2}{c}{{Medium}} & \multicolumn{2}{c}{{Hard}} \\
\cmidrule(lr){2-3} \cmidrule(lr){4-5} \cmidrule(lr){6-7} 
& gIoU & cIoU &gIoU & cIoU & gIoU & cIoU \\ 
% \midrule
        \cmidrule(lr){1-1} \cmidrule(lr){2-3} \cmidrule(lr){4-5} \cmidrule(lr){6-7}   
Baseline-7B &67.7&66.4&51.2&50.2&47.0&46.3\\
+ pre-reasoning &67.3&66.7&57.2&\textbf{58.1}&57.0&\textbf{58.3}\\
+ inter-reasoning &64.3&64.7&55.5&56.3&53.9&54.8\\
+ selective reasoning &\textbf{68.9}&\textbf{67.2}&\textbf{58.9}&{57.2}&\textbf{57.2}&57.7\\
% \midrule
        \cmidrule(lr){1-1} \cmidrule(lr){2-3} \cmidrule(lr){4-5} \cmidrule(lr){6-7} 
Baseline-13B &69.2&70.3&55.2&56.2&51.7&52.2\\
+ pre-reasoning &69.7&69.3&62.7&62.2&61.9&61.2\\
+ inter-reasoning &67.2&68.1&60.9&59.2&58.2&57.2\\
+ selective reasoning &\textbf{70.8}&\textbf{71.3}&\textbf{64.2}&\textbf{65.2}&\textbf{62.9}&\textbf{63.8}\\
% \midrule
% Grounded-Align-7B &68.2&67.3&49.3&46.9&59.0&60.1  \\
% Grounded-Align-13B &73.7&68.6&55.2&50.1&63.9&66.8  \\
\bottomrule
\end{tabular}
\label{tab:reasonseg}
    \vspace{-1.3em}
\end{table}

\begin{table}[t]
\centering
\caption{Performance comparison on the REC and RES tasks.}
% \resizebox{0.5\textwidth}{!}{
    \footnotesize
    % \vspace{-0.05cm}
       \vspace{0.1em}
    \setlength\tabcolsep{0.27cm}
\begin{tabular}{l|cc|cc|cc}
% \begin{tabular}{l|cc|cc|cc}
\toprule
\multirow{2}{*}{Models} & \multicolumn{2}{c|}{RefCOCO} & \multicolumn{2}{c|}{RefCOCO+} & \multicolumn{2}{c}{RefCOCOg}\\
&REC&RES&REC&RES &REC&RES\\
% \midrule
        \cmidrule(lr){1-1} \cmidrule(lr){2-3} \cmidrule(lr){4-5} \cmidrule(lr){6-7} 
Kosmos-2~\cite{peng2023kosmos} &52.3 & --& 45.4&--  & 60.5& --  \\
LISA-7B (ft)~\cite{lai2024lisa}  &--&74.9&-- &65.1&-- &67.9 \\
LLaVASeg-7B (ft)~\cite{yang2024empowering} &--&76.2&-- &65.7&-- &69.8 \\
MiniGPT v2-7B~\cite{chen2023minigpt} & 88.0& -- & 79.5& -- & 84.1& --  \\
Shikra-7B~\cite{chen2023shikra} & 87.0 & --   & 81.6 & --   & 82.2 & --     \\
Ferret-7B~\cite{you2023ferret} & 87.4 & --  & 80.7 & --   & 83.9 & --  \\
LLaVA-Grounding-7B~\cite{zhang2024llava} & 89.1   & {77.1} & {\textbf{81.6}}  & 68.7 & {84.8}   & {71.5}   \\
VisionLLM v2~\cite{wu2025visionllm} &87.9&76.6&77.6&64.5&82.9&70.7\\
PixelLLM~\cite{ren2024pixellm} &--&73.0&--&66.3&--&69.3 \\
GLaMM~\cite{rasheed2024glamm} &--&79.5&--&\textbf{72.6}&--&74.2\\
GSVA-7B (ft)~\cite{xia2024gsva} &86.2&77.2&72.8&65.9&81.5&72.7\\
% \midrule
        \cmidrule(lr){1-1} \cmidrule(lr){2-3} \cmidrule(lr){4-5} \cmidrule(lr){6-7} 
MMGrounded-PostAlign-7B   &88.2&77.9&78.4&68.2&83.3&73.2 \\
MMGrounded-PostAlign-13B  &\textbf{89.2}&\textbf{79.7}&80.1&{70.9}&\textbf{85.3}&\textbf{74.8 }\\
\bottomrule
\end{tabular}
\label{tab:refcoco}
\vspace{-1.3em}
\end{table}

\noindent \textbf{Finding 5: Our Architecture Provides a Potential Approach for Visual Grounding with MLLMs.}
We also evaluate our method’s visual grounding performance on the RefCOCO series and ReasonSeg benchmarks. The results surprisingly show that our framework not only improves the MLLM’s visual understanding through the visual grounding module, but also that the MLLM can assist the visual grounding module in enhancing its zero-shot grounding capabilities. This suggests a potential approach for visual grounding empowered by MLLMs.
However, it is important to emphasize that the primary goal of our work is not to optimize for visual grounding tasks directly. Instead, we aim to leverage multimodal grounding (including both visual and text grounding) as evidence to suppress hallucinations in MLLMs and improve their visual understanding. Consequently, we do not increase the amount of grounding-specific training data to further optimize performance on grounding benchmarks.
In Table~\ref{tab:refcoco}, for the REC and RES tasks, our approach achieves competitive results compared to specialized MLLM-based grounding models. This highlights the superior pixel-level grounding capabilities of our method. 
% Additional results on other visual grounding tasks can be found in the \textit{Appendices}.

\begin{wrapfigure}{r}{0.44\textwidth}
  \centering
  \includegraphics[width=0.44\textwidth]{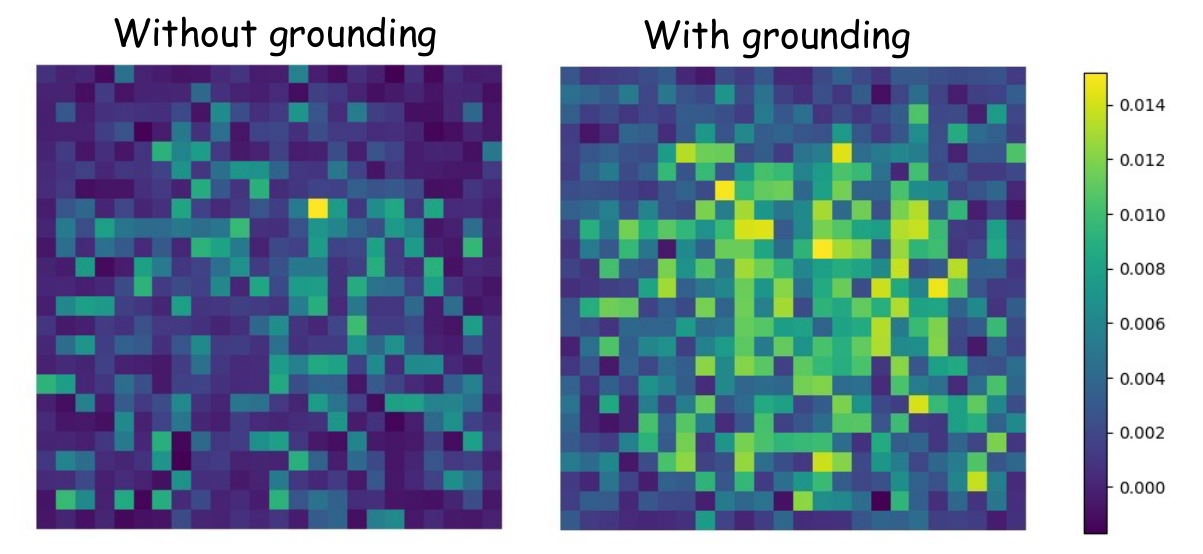}
  \caption{Average attention to image features without (left) and with grounding (right).}
  % Without grounding, attention is sparse and scattered, showing weak alignment with relevant visual regions. With grounding, attention is more concentrated over meaningful areas, enhancing alignment between textual outputs and visual features for more coherent responses.}
  \label{fig:vis}
  \vspace{-1.4em}
\end{wrapfigure}
\subsection{Qualitative Analysis}
To investigate how grounding influences the model's attention to image features, we visualize the average attention weights assigned to image features over 500 MSCOCO samples during response generation under two conditions: without grounding and with grounding.
To emphasize the most salient attention patterns and reduce noise, we apply Top-k Mean Pooling at three levels: For each output token, we select the top-3 highest attention values across all hidden layers and compute their mean. We then aggregate the top-3 attention values across all self-attention heads for each token. Finally, to ensure consistency between the two conditions, we average the top-l attention values across output tokens, where l is the minimum number of tokens generated between the without grounding and with grounding cases.
The aggregated attention maps in Figure~\ref{fig:vis} show a clear distinction between the two conditions. The without grounding attention map exhibits sparse and scattered attention. In contrast, the with grounding map shows significantly more concentrated attention, particularly over relevant image areas. This suggests that grounding effectively guides the model to align textual outputs with pertinent visual features, resulting in more visually grounded responses.

\section{Conclusion}
In this paper, we propose MMGrounded-PostAlign, a post-multimodal alignment framework designed to enhance the visual understanding capabilities of MLLMs. Our approach addresses the challenges of over-reliance on spurious correlations by integrating a grounding module that enables both visual and textual grounding. This ensures that the model’s outputs are firmly anchored in actual visual and textual evidence.
We propose a negative rejection mechanism within the visual grounding module to mitigate hallucinations caused by linguistic biases, helping the model distinguish grounded objects from non-existent ones. 
Additionally, our selective reasoning in textual grounding adapts the model’s reasoning strategy based on query complexity, allowing for more accurate and contextually relevant responses for tasks of varying complexity levels.
Extensive evaluations on POPE, HaloQuest, MME, MMBench, VQAv2, ReasonSeg, and RefCOCO demonstrate the effectiveness of our approach in improving visual understanding, reducing hallucinations, and enhancing grounding accuracy.

% \section*{References}

% {
%     \small
%     % \bibliographystyle{ieeenat_fullname}
%     \bibliography{main}
% }

% \bibliographystyle{plainnat}
\bibliographystyle{unsrt}
\bibliography{main}

% References follow the acknowledgments in the camera-ready paper. Use unnumbered first-level heading for
% the references. Any choice of citation style is acceptable as long as you are
% consistent. It is permissible to reduce the font size to \verb+small+ (9 point)
% when listing the references.
% Note that the Reference section does not count towards the page limit.
\medskip

\end{document}